\documentclass{article}
\usepackage{ijcai18}

\usepackage{times}
\usepackage{microtype}
\usepackage{bm}
\usepackage{xcolor}
\usepackage{soul}
\usepackage{graphicx}

\usepackage[utf8]{inputenc}
\usepackage[small]{caption}
\usepackage[boxed, algoruled, vlined, linesnumbered]{algorithm2e} 
\usepackage{url}
\SetKwInOut{Initialize}{Initialize}
\newcommand\bmu[1]{\color{blue}{\bm{#1}}}
\newcommand\bmg[1]{\color{black}{\bm{#1}}}

\usepackage{amsmath,amssymb}
\usepackage{listings}
\title{Back to Basics: Benchmarking Canonical Evolution Strategies for Playing Atari}

\author{
Patryk Chrabaszcz, 
Ilya Loshchilov, 
Frank Hutter 
\\ 
University of Freiburg, Freiburg, Germany \\
\{chrabasp,ilya,fh\}@cs.uni-freiburg.de
}

\newcommand{\note}[1]{}
\renewcommand{\note}[1]{~\\\frame{\begin{minipage}[c]{0.49\textwidth}\vspace{2pt}\center{#1}\vspace{2pt}\end{minipage}}\vspace{3pt}\\}

\newcommand{\citet}[1]{\citeauthor{#1} [\citeyear{#1}]}

\begin{document}

\maketitle

\begin{abstract}
Evolution Strategies (ES) have recently been demonstrated to be a viable alternative to reinforcement learning (RL) algorithms on a set of challenging deep RL problems, including Atari games and MuJoCo humanoid locomotion benchmarks. While the ES algorithms in that work belonged to the specialized class of natural evolution strategies (which resemble approximate gradient RL algorithms, such as REINFORCE), we demonstrate that even a very basic canonical ES algorithm can achieve the same or even better performance. This success of a basic ES algorithm suggests that the state-of-the-art can be advanced further by integrating the many advances made in the field of ES in the last decades.

We also demonstrate qualitatively that ES algorithms have very different performance characteristics than traditional RL algorithms: on some games, they learn to exploit the environment and perform much better while on others they can get stuck in suboptimal local minima. Combining their strengths with those of traditional RL algorithms is therefore likely to lead to new advances in the state of the art.
  
\end{abstract}
\section{Introduction}
    {In machine learning, Evolution Strategies (ES) are mainly used for direct policy search in reinforcement learning \cite{gomez2008accelerated,heidrich2009hoeffding,stulp2013robot,salimans2017evolution} and hyperparameter tuning in supervised learning, e.g., for Support Vector Machines \cite{glasmachers2008uncertainty,igel2011evolutionary} and Deep Neural Networks \cite{loshchilov2016cma}}. 
 
Recently it has been shown~\cite{salimans2017evolution} that ES algorithms can be used for tasks which are dominated by deep reinforcement learning (RL) algorithms. Those tasks include learning a policy with discrete action set to control an agent's behavior in a wide variety of Atari environments, as well as learning a policy with continuous action space for agents operating in MuJoCo~\cite{todorov2012mujoco} environments. ES algorithms offer a set of attractive advantages when compared to deep RL algorithms:
\begin{itemize}
  \item They are highly parallelizable, since the amount of information that has to be exchanged between workers does not depend on the network size.
  \item Depending on the problem, they can offer better exploration, and as a result different training runs can converge to qualitatively different solutions. 
  \item They are not sensitive to the distribution of rewards and do not require careful tuning of discount factors while still facilitating long-term foresight more than traditional discount-based RL algorithms.
    \item They can be used for the optimization of non-differentiable policy functions.
\end{itemize}

In this work, we go one step further than \citet{salimans2017evolution} and study the applicability of even simpler ES algorithm to the task of learning a policy network for playing Atari games. \citet{salimans2017evolution} used a specialized ES algorithm that belongs to the class of Natural Evolution Strategies (NES)~\cite{wierstra2008natural}, which computes approximate gradients similar to the REINFORCE algorithm~\cite{williams1992simple}. Here, we demonstrate that very comparable results can already be achieved with a simpler very basic Canonical ES algorithm from the 1970s~\cite{rechenberg1973evolutionsstrategie,rudolph1997convergence}.

Our contributions in this work are as follows:
\begin{itemize}
	\item We demonstrate that even a very basic Canonical ES algorithm is able to match (and sometimes supersede) the performance of the Natural Evolution Strategy used by \citet{salimans2017evolution} for playing Atari games.

\item We demonstrate that after 5 hours of training, Canonical ES is able to find novel solutions that exploit the game design and even find bugs in one game that allow them to achieve unprecedented high scores.

\item We experimentally study the performance characteristics of both ES algorithms, demonstrating that (1) individual runs have high variance in performance and that (2) longer runs (5h instead of 1h) lead to significant further performance improvements.

\item By demonstrating that Canonical ES is a competitive alternative to traditional RL algorithms and the specialized ES algorithms tested so far on the Atari domain we set a benchmark for future work on modern ES variants that are directly based on the canonical version. 
\end{itemize}

\section{Background}
	In this section, we discuss background on RL for playing Atari and on the previously-introduced NES method.

\subsection{Reinforcement Learning for Playing Atari}
	In the Atari task of the OpenAI gym environment~\cite{brockman2016openai}, the agent needs to learn to maximize its cumulative reward solely by interacting with the environment (i.e., playing the game). Inputs to the agent include the raw pixels displayed on the Atari console as well as the reward signal; its actions correspond to joystick movements to execute. 

Recent developments in the field of deep RL have made it possible to address challenging problems that require processing of high dimensional inputs, such as the raw images in this Atari domain, which can be adressed by deep convolutional neural networks. This approach was popularized by Google DeepMind's Nature paper on the deep Q network (DQN)~\cite{mnih2015human}, a Q-learning method that estimates the utility of an action given a current state by means of a deep neural network. Given this network for approximating the Q function, in any state $s$, DQN's policy then simply selects the action $a$ with the largest predicted Q value $Q(s,a)$. 

	While DQN requires this maximization over the action space, policy gradient algorithms directly parameterize a policy networks that maps a state to a probability distribution over actions. Policy gradient algorithms, such as the Asynchronous Advantage Actor Critic (A3C)~\cite{mnih2016asynchronous}, directly optimize this policy network.

\begin{figure}[t]
\centering
\includegraphics[width=0.45\textwidth]{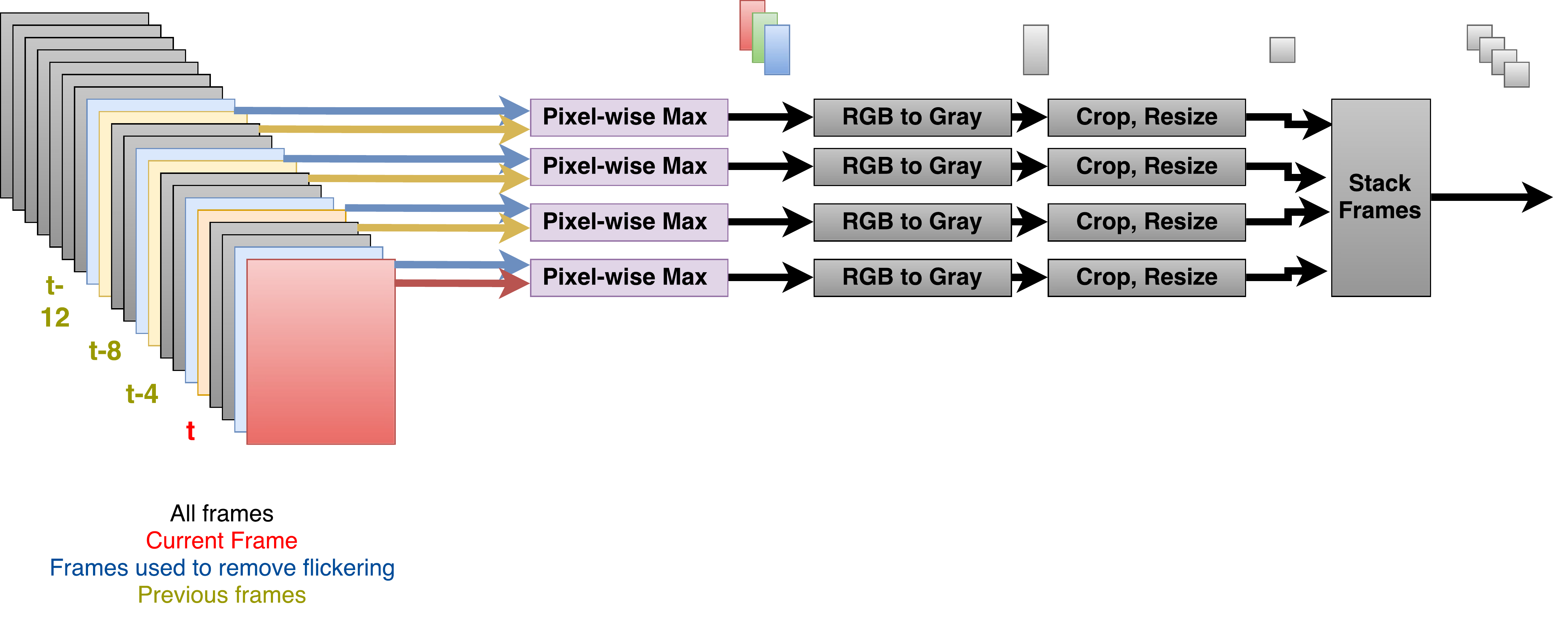}
\vspace*{-0.2cm}
\caption{\label{fig:Preprocessing}Preprocessing pipeline. Take every 4th frame, apply max operation to remove screen flickering, convert to grayscale, resize/crop, stack 4 last frames.}
\end{figure}

\begin{figure}[t]
\centering
\includegraphics[width=0.09\textwidth]{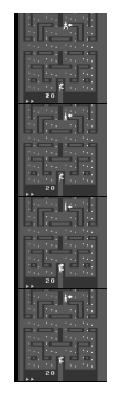}
\includegraphics[width=0.09\textwidth]{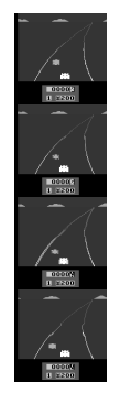}
\includegraphics[width=0.09\textwidth]{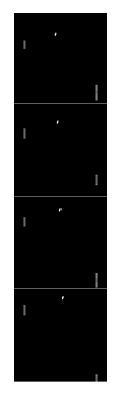}
\includegraphics[width=0.09\textwidth]{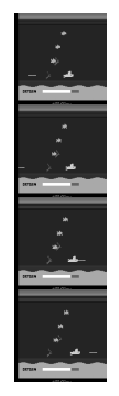}
\includegraphics[width=0.09\textwidth]{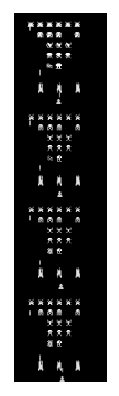}
\vspace*{-0.2cm}
\caption{\label{fig:GameScreens}State representation (84x84x4 tensor) for 5 different games: Alien, Enduro, Pong, Seaquest and SpaceInvaders. Channels are shown on top of each other for better visualization.}
\end{figure}

\vspace*{-0.5cm}
\paragraph{State representation.} In Atari games it is important to model the state to include information from previous frames that will influence an agent's performance. In this work we use the following standard preprocessing pipeline~\cite{mnih2015human} with an implementation provided by OpenAI~\cite{baselines}. First, we apply preprocessing to the screen frames to reduce the dimensionality and remove artifacts related to the technical limitations of the Atari game console (flickering). Specifically, we apply a pixel-wise max operation on the current frame and the one preceeding it. Next, we convert the output from this operation into a grayscale image, resize it and crop it to 84x84 pixels. At the end of this pipeline, we stack together the result of the 4 last frames produced this way to construct a 84x84x4 state tensor. Also following common practice, to speed up policy evaluation (and thus reduce training time), instead of collecting every frame and making a decision at every step, we collect every 4th frame (3rd frame for SpaceInvaders to make the laser visible~\cite{mnih2013playing}) and apply the same action for all frames in between. Figure \ref{fig:Preprocessing} visualizes the full preprocessing pipeline. Figure \ref{fig:GameScreens} shows an example of the state representation for 5 different games. 
    
\subsection{Natural Evolution for Playing Atari}

\begin{algorithm}[t] 
\footnotesize
\caption{OpenAI ES\label{alg:OpenAI_ES}}
 \KwIn{\\
    $optimizer$ - Optimizer function \\
    $\sigma$ - Mutation step-size \\
    $\lambda$ - Population size \\
    $\theta_{0}$ - Initial policy parameters \\
    $F$ - Policy evaluation function \\
 }
 \For{t = 0, 1, ...}{
  \For{i = 1, 2, ... $\frac{\lambda}{2}$}{
    Sample noise vector: $\epsilon_{i} \sim \mathcal{N}(0, I)$ \\
    Evaluate score in the game: $s^{+}_{i} \leftarrow F(\theta_{t} + \sigma*\epsilon_{i})$ \\
    Evaluate score in the game: $s^{-}_{i} \leftarrow F(\theta_{t} - \sigma*\epsilon_{i})$ \\
  }
  Compute normalized ranks: $r = ranks(s), r_{i} \in [0, 1)$\\ 
  Estimate gradient: $g \leftarrow \frac{1}{\sigma * \lambda} \sum_{i=1}^{\lambda} (r_{i}* \epsilon_{i})$ \\
  Update policy network: $\theta_{t+1} \leftarrow \theta_{t} + optimizer(g)$ \\
 }
\end{algorithm}

	\citet{salimans2017evolution} recently demonstrated that an ES algorithm from the specialized class of Natural Evolution Strategies (NES; \citet{wierstra2008natural}) can be used to successfully train policy networks in a set of RL benchmark environments (Atari, MuJoCo) and compete with state-of-the-art RL algorithms. Algorithm \ref{alg:OpenAI_ES} describes their approach on a high level. In a nutshell, it evolves a distribution over policy networks over time by evaluating a population of $\lambda$ different networks in each iteration, starting from initial policy parameter vector $\theta_0$. At each iteration $t$, the algorithm evaluates the game scores $F(\cdot)$ of $\lambda$ different policy parameter vectors centered around $\theta_t$ (lines 3-5) to estimate a gradient signal, using mirrored sampling~\cite{brockhoff2010mirrored} to reduce the variance of this estimate. Since the $\lambda$ game evaluations are independent of each other, ES can make very efficient use of parallel compute resources. The resulting $\lambda$ game scores are then ranked (line 6), making the algorithm invariant to their scale; as noted by the authors, this approach (called fitness shaping~\cite{wierstra2014natural} but used in all ESs since the 1970s) 
decreases the probability of falling into local optima early and lowers the influence of outliers. Based on these $\lambda$ ranks of local steps around $\theta_t$, the algorithm approximates a gradient $g$ (line 7) and uses this with a modern version of gradient descent (Adam~\cite{kingma2014adam} or SGD with momentum)
with weight decay to compute a robust parameter update step (line 8) in order to move $\theta_t$ towards the parameter vectors that achieved higher scores.
   
We note that the computation of the approximate gradient $g$ in line 7 follows the same approach as the well-known policy gradient algorithm REINFORCE. This can be shown as follows. Denoting the distribution from which we draw policy network parameters $\theta$ as $p_{\psi}$,
the gradient of the expected reward $F(\theta)$ with respect to $\psi$ is:
\begin{equation}
	\nabla_{\psi} \mathbb{E}_{\theta \sim p_{\psi}}\{F(\theta)\} = \mathbb{E}_{\theta \sim p_{\psi}}\{F(\theta)\nabla_{\psi}\log{p_{\psi}(\theta)}\}. 
\end{equation}

	Because $p_{\psi}$ is chosen as an isotropic Gaussian distribution with mean $\theta_t$ and fixed standard deviation (mutation step-size) $\sigma$, the only parameter of $p_{\psi}$ is $\theta_t$ and we have:
\begin{equation}
	\nabla_{\psi}\log{p_{\psi}(\theta)} = \nabla_{\theta_t}\log\frac{1}{{\sigma \sqrt {2\pi } }}e^{{{ - \left( {\theta - \theta_t } \right)^2 } \mathord{\left/ {\vphantom {{ - \left( {x - \theta_t } \right)^2 } {2\sigma ^2 }}} \right. \kern-\nulldelimiterspace} {2\sigma ^2 }}} = \frac{\theta-\theta_t}{\sigma^{2}}
\end{equation}
\noindent{}and therefore the following identity holds:
\begin{eqnarray}
	\nabla_{\psi} \mathbb{E}_{\theta \sim p_{\psi}}\{F(\theta)\}\!\!\!\!\!&=&\!\!\!\!\! \mathbb{E}_{\theta \sim p_{\psi}}\{F(\theta) * \frac{\theta-\theta_t}{\sigma^{2}}\}\\
     &\approx&\!\!\!\!\! \frac{1}{\sigma*\lambda} \sum_{i=1}^{\lambda} F(\theta^{(i)}) * \frac{(\theta^{(i)}- \theta_t)}{\sigma},\label{eq:like_line7} 
\end{eqnarray}\noindent{}where the last step is simply an approximation by $\lambda$ samples $\theta^{(i)} \sim p_{\psi}$. Equation \ref{eq:like_line7} is exactly as in line 7 of the algorithm except that the raw game scores $F(\theta^{(i)})$ are replaced with their ranks $r_i$ due to fitness shaping.

\citet{salimans2017evolution} also made two further contributions to stabilize the training and improve performance.
Firstly, they introduced a novel parallelization technique (which uses a noise table to reduce communication cost in order to scale to a large number of $\lambda$ parallel workers) and
used virtual batch normalization~\cite{salimans2016improved} to make the network output more sensitive to the noise in the parameter space.

\section{Canonical ES}
	While the specialized ES algorithm proposed by OpenAI is equivalent to a policy gradient algorithm, in this work we consider a very basic canonical ES algorithm that belongs to the prominent family of $(\mu, \lambda)-ES$ optimization algorithms. Algorithm \ref{algo:Canonical_ES} illustrates this simple approach. Starting from a random parameter vector $\theta_{0}$, in each iteration $t$ we generate an offspring population of size $\lambda$. For each element of the population, we add sampled mutation noise $\epsilon_{i} \sim \mathcal{N}(0, \sigma^2)$ to the current parameter vector $\theta_{t}$ (line 3) and evaluate the game score of the resulting vector by one episode rollout (line 4). We then pick the top $\mu$ parameter vectors according to the collected score and form a new parameter vector $\theta_{t+1}$ as their weighted mean (lines 5-6). 
    
    This algorithm is very basic in its setting: we do not use mirrored sampling, we do not decay the parameters, we do not use any advanced optimizer. The standard weights used to compute the weighted mean of the top $\mu$ solutions fulfill a similar function to the fitness shaping implemented in OpenAI ES. 
    The new elements introduced by \citet{salimans2017evolution} that we \emph{do} use are virtual batch normalization (which is a component of the game evaluations $F$ and not really of the ES algorithm itself) and the efficient parallelization of ES using a random noise table. 

   We initially implemented the Cumulative Step-size $\sigma$ Adaptation (CSA) procedure~\cite{hansen1996adapting}, which is standard in canonical ES algorithms. However, due to the high time cost of game evaluations, during our time-limited training, we are only able to perform up to thousands update iterations. Since the dimensionality of the parameter vector is relatively large (1.7M), this results in only a negligible change of $\sigma$ during the training. Therefore, effectively, our algorithm used a fixed step-size and thus we removed step-size adaptation from the description from Algorithm 2 making it even somewhat simpler than typical ES. We employed weighted recombination \cite{rudolph1997convergence} and weights $w$ as in CSA-ES.

\begin{algorithm} 
\footnotesize
\caption{Canonical ES Algorithm\label{algo:Canonical_ES}}
 \KwIn{\\
    $\sigma$ - Mutation step-size \\
    $\theta_{0}$ - Initial policy parameters \\
    $F$ - Policy evaluation function \\
    $\lambda$ - Offspring population size \\ 
    $\mu$ - Parent population size \\
 }
 \Initialize{ \\
   \vspace*{0.1cm}\hspace*{-1.5cm}$w_{i} = \frac{\log(\mu + 0.5) - \log(i)}{\sum_{j=1}^{\mu}\log(\mu + 0.5) - \log(j)}$ \\
}
 \For{$t = 0, 1, ...$}{
  \For{$i = 1 ... \lambda$}{
      Sample noise: $\epsilon_{i} \sim \mathcal{N}(0, I)$ \\
      Evaluate score in the game: $s_{i} \leftarrow F(\theta_{t} + \sigma*\epsilon_{i})$ \\
	}
    Sort $(\epsilon_1, \dots, \epsilon_\lambda)$ according to $s$ ($\epsilon_i$ with best $s_i$ first)\\
    Update policy: $\theta_{t+1} \leftarrow \theta_{t} + \sigma*\sum_{j=1}^{\mu}w_{j} * \epsilon_{j}$\\
    Optionally, update step size $\sigma$ (see text)
 }

\end{algorithm}
\section{Experiments}

	In our experiments, we evaluate the performance of the Canonical ES on a subset of 8 Atari games available in OpenAI Gym ~\cite{brockman2016openai}. We selected these games to represent different levels of difficulty, ranging from simple ones like Pong and Breakout to complex games like Qbert and Alien. We make our implementation of the Canonical ES algorithm available online at \url{https://github.com/PatrykChrabaszcz/Canonical_ES_Atari}.
    
\begin{figure}[t]
\centering
\includegraphics[width=0.45\textwidth]{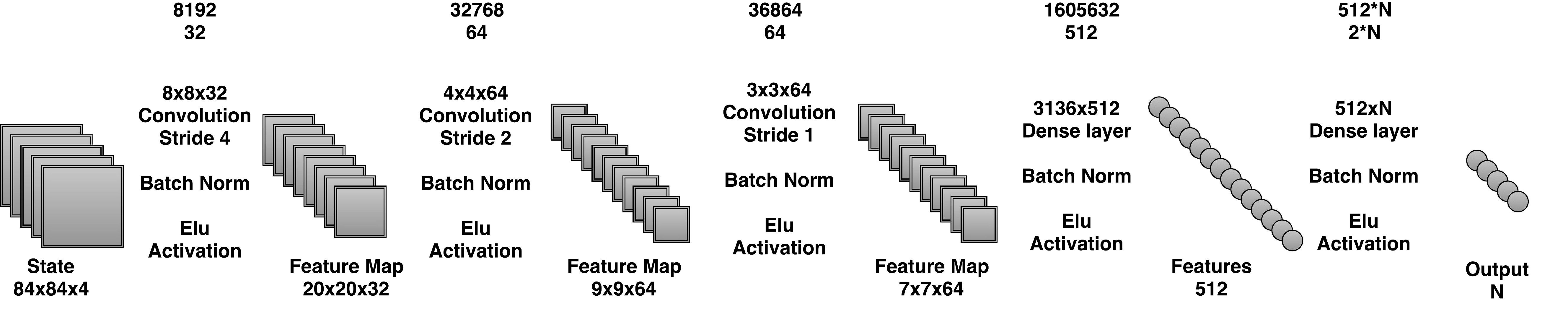}
\caption{\label{fig:Architecture}Neural network architecture. Numbers on top show number of parameters in each layer (kernel parameters and batch norm parameters). Each batch norm layer has a trainable shift parameter $\beta$; the last batch norm has an additional trainable scale parameter $\alpha$.}
\end{figure}

    We compare our results against those obtained with the ES algorithm proposed by OpenAI~\cite{salimans2017evolution}. Since no implementation of that algorithm is publicly available for Atari games, we re-implemented it with help from OpenAI\footnote{We thank Tim Salimans for his helpful email support.} and the results of our implementation (which we refer to as ``OpenAI ES (our)'') roughly match those reported by~\citet{salimans2017evolution} (see Table \ref{table:Scores}).  
\vspace*{-0.2cm}
\paragraph{{Network Architecture.}} 
We use the same network structure as the original DQN work~\cite{mnih2015human}, only changing the activation function from ReLU to ELU~\cite{clevert2015fast} and adding batch normalization layers~\cite{ioffe2015batch}. The network as presented in Figure \ref{fig:Architecture} has approximately 1.7M parameters. We initialize network weights using samples from a normal distribution $\mathcal{N}(\mu=0, \sigma=0.05)$. 

\vspace*{-0.5cm}
    \paragraph{{Virtual Batch Normalization.}}Following \citet{salimans2017evolution}, we use virtual batch normalization~\cite{salimans2016improved}.
    In order to collect the reference batch, at the beginning of the training we play the game using random actions. In each step, we save the corresponding state with the probability $p(save) = 1\%$ and stop when 128 samples have been collected. 
\vspace*{-0.5cm}
    \paragraph{{Training.}}For each game and each ES variant we tested, we performed 3 training runs, each on 400 CPUs with a time budget of 10 hours. Every worker (CPU) evaluates 2 offspring solutions, meaning that our setting is roughly the same as training for 5 hours with full parallelization (800 CPUs); therefore, we label this setting as ``5 hours''. In addition, we save the solution proposed after 2 hours of training (equivalent to 1 hour with full parallelization) or after 1 billion training frames (whatever comes first) to allow for a fair comparison with results reported by \citet{salimans2017evolution}; we label this setting as ``1 hour''. During training, one CPU is reserved to evaluate the performance of the solution proposed in the current iteration; hence, the offspring population size in our experiments is $\lambda=798$.
    In each decision step, the agent passes its current environment state through the network and performs an action that corresponds to the output with the highest value. 
    We limit episodes to have a maximum length of 25k steps; we do not adjust this value during training.

An episode includes multiple lives, and we do not terminate an episode after the agent dies the first time in order to allow the learning of strategies that span across multiple lives. We start each episode with up to 30 initial random no-op actions. 
    
\begin{figure}
\centering
\includegraphics[width=0.45\textwidth]{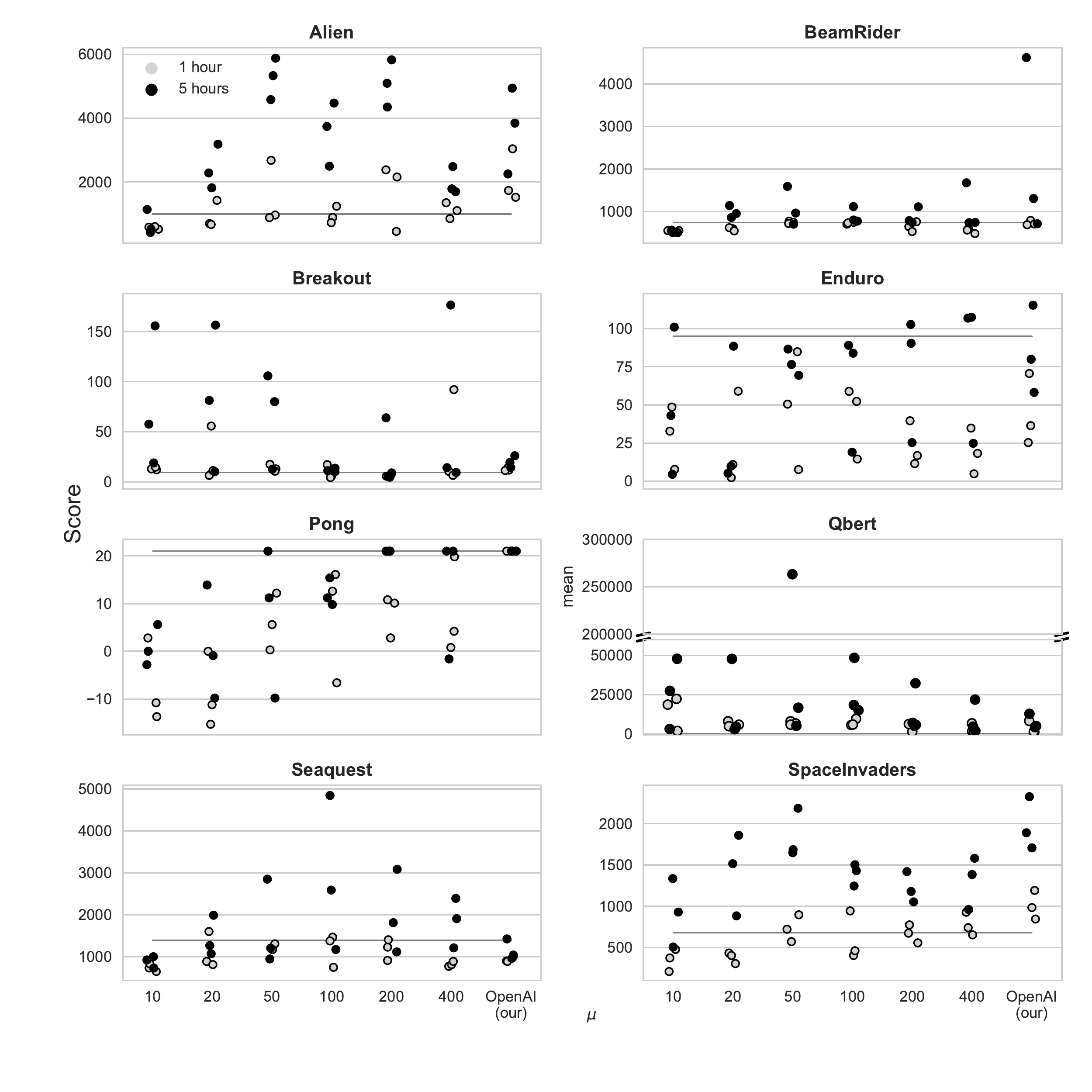}
\vspace*{-0.7cm}
\caption{\label{fig:Scores} Final evaluation scores (mean across 30 evaluation rollouts with random initial no-ops) for Canonical ES ($\mu \in {10, 20, 50, 100, 200, 400}$) and for our implementation of OpenAI ES. For each setting we use population size $ \lambda=798$ and report the performance from 3 separate training runs. We report evaluation scores after 1 hour and after 5 hours of training. The horizontal line indicates the results reported by \protect\citet{salimans2017evolution}.}
\end{figure}

\begin{figure}[h]
\centering
\includegraphics[width=0.45\textwidth]{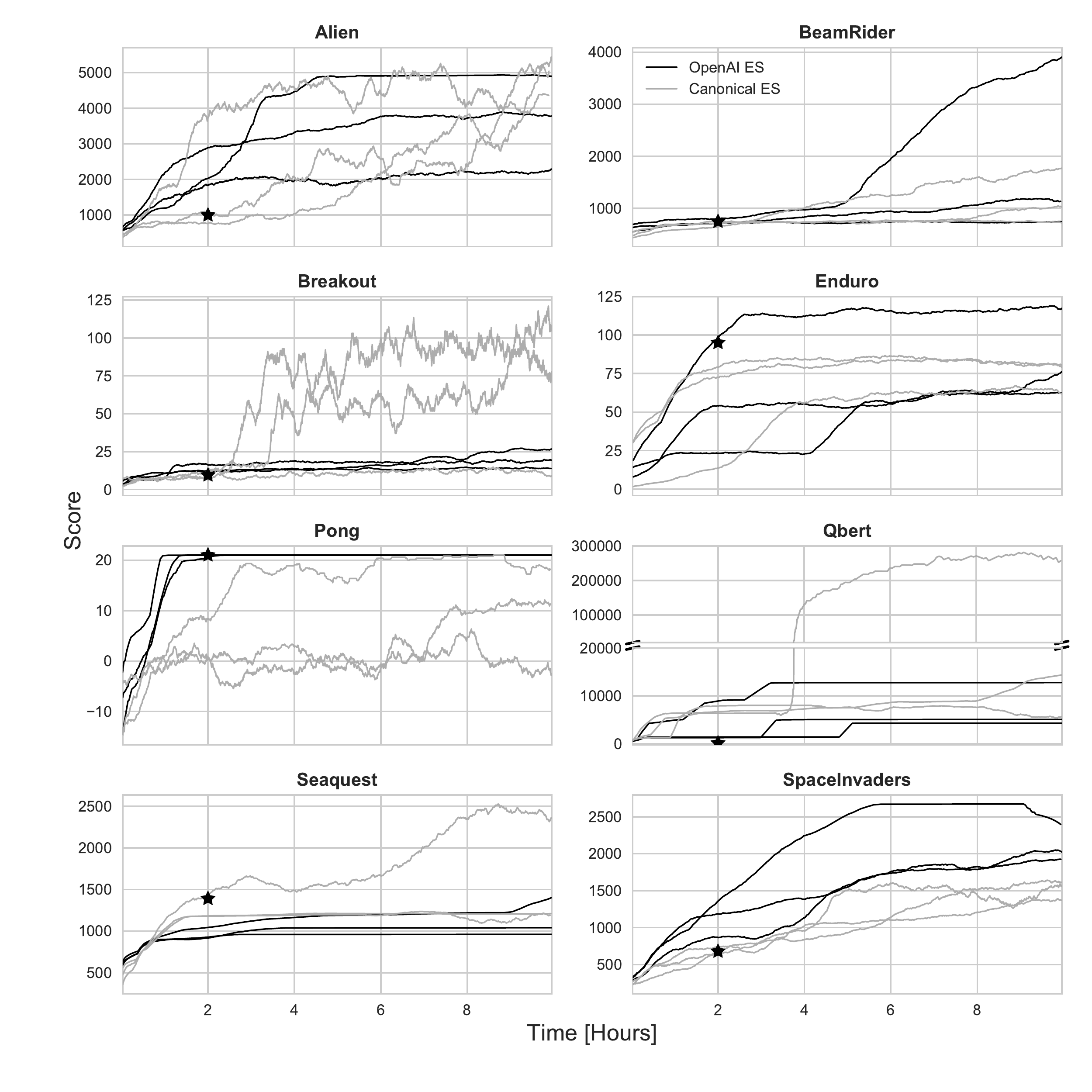}
\vspace*{-0.5cm}
\caption{\label{fig:TimePlot}{Training curves for CanonicalES ($\mu=50$) and OpenAI ES (our). At each iteration $t$ we evaluate currently proposed solution $\theta{t}$ two times using one CPU, because of that values reported in the Table\ref{table:Scores} (mean over 30 evaluation runs) might differ. For better plot quality we filter out the noise. We observe that both algorithms often get stuck in local optima.}}
\end{figure}

\begin{table*}[t!]
\footnotesize
\centering
\begin{tabular}{l | c | l l | l l}
 \hline
   				& OpenAI ES		    & OpenAI ES (our) 		& Canonical ES				& OpenAI ES (our)         & Canonical ES 		        \\ [0.5ex]
        		& 1 hour		    & 1 hour	 		    & 1 hour 					& 5 hours	              & 5 hours	   		            \\ \hline
 Alien			& 			        & $\bmg{3040\pm276.8}$	& $2679.3\pm1477.3$   		& $4940\pm0$              & $\bmu{5878.7\pm1724.7}$		\\
 Alien			& $994$ 	    	& $\bmu{1733.7\pm493.2}$& $965.3\pm229.8$	        & $3843.3\pm228.7$ 	      & $\bmu{5331.3\pm990.1}$		\\
 Alien			&  				    & $\bmu{1522.3\pm790.3}$& $885\pm469.1$             & $2253\pm769.4$		  & $\bmu{4581.3\pm299.1}$		\\ \hline

 BeamRider		&  				    & $\bmg{792.3\pm146.6}$	& $774.5\pm202.7$ 			& $\bmu{4617.1\pm1173.3}$ & $1591.3\pm575.5$		    \\
 BeamRider		& $744$ 	    	& $708.3\pm194.7$		& $\bmg{746.9\pm197.8}$ 		& $\bmu{1305.9\pm450.4}$  & $965.3\pm441.4$		        \\
 BeamRider		&  				    & $690.7\pm87.7$		& $\bmu{719.6\pm197.4}$ 	& $\bm{714.3\pm189.9}$    & $703.5\pm159.8$		        \\ \hline

 Breakout		&  				    & $14.3\pm6.5$		    & $\bmu{17.5\pm19.4}$ 		& $26.1\pm5.8$		      & $\bmu{105.7\pm158}$	        \\
 Breakout		& $9.5$ 	    	& $11.8\pm3.3$	    	& $\bmu{13\pm17.1}$ 		& $19.4\pm6.6$  	      & $\bmu{80\pm143.4}$	        \\
 Breakout		&  				    & $\bmu{11.4\pm3.6}$ 	& $10.7\pm15.1$ 			& $\bmu{14.2\pm2.7}$   	  & $12.7\pm17.7$		        \\ \hline

 Enduro			&  				    & $70.6\pm17.2$		    & $\bmu{84.9\pm22.3}$ 		& $\bmu{115.4\pm16.6}$    & $86.6\pm19.1$		        \\
 Enduro			& $95$ 	    		& $36.4\pm12.4$		    & $\bmu{50.5\pm15.3}$ 		& $\bm{79.9\pm18}$		  & $76.5\pm17.7$		        \\
 Enduro			& 				    & $\bmu{25.3\pm9.6}$	& $7.6\pm5.1$ 				& $58.2\pm10.5$  	      & $\bm{69.4\pm32.8}$		    \\ \hline

 Pong			&  				    & $\bmu{21.0\pm0.0}$&   $12.2\pm16.6$ 			    & $\bmg{21.0\pm0.0}$       & $\bmg{21.0\pm0.0}$		    \\
 Pong			& $21$ 	    		& $\bmu{21.0\pm0.0}$	& $5.6\pm20.2$ 	            & $\bmu{21\pm0}$          & $11.2\pm17.8$		        \\
 Pong			&  				    & $\bmu{21.0\pm0.0}$	& $0.3\pm20.7$ 			    & $\bmu{21\pm0}$          & $-9.8\pm18.6$		        \\ \hline

 Qbert			&  				    & $\bmu{8275\pm0}$		& $8000\pm0$ 			    & $12775\pm0$    	      & $\bmu{263242\pm433050}$      \\
 Qbert			& $147.5$		    & $1400\pm0$		    & $\bmu{6625\pm0}$ 			& $5075\pm0$    	      & $\bmu{16673.3\pm6.2}$		\\
 Qbert			&  				    & $1250\pm0$		    & $\bmu{5850\pm0}$ 			& $4300\pm0$     	      & $\bmg{5136.7\pm4093.9}$		\\ \hline

 Seaquest		&  				    & $1006\pm20.1$		    & $\bmu{1306.7\pm262.7}$ 	& $1424\pm26.5$    	      & $\bmu{2849.7\pm599.4}$		\\
 Seaquest		& $1390$		    & $898\pm31.6$		    & $\bmu{1188\pm24}$ 		& $1040\pm0$    	      & $\bmu{1202.7\pm27.2}$		\\
 Seaquest		& 				    & $887.3\pm20.3$	    & $\bmu{1170.7\pm23.5}$		& $\bmg{960\pm0}$     	  & $946.7\pm275.1$		        \\ \hline

 SpaceInvaders	&  				    & $\bmu{1191.3\pm84.6}$	& $896.7\pm123$ 		    & $\bmg{2326.5\pm547.6}$   & $2186\pm1278.8$		        \\
 SpaceInvaders	& 678.5		 	    & $\bmu{983.7\pm158.5}$	& $721.5\pm115$ 		    & $\bmg{1889.3\pm294.3}$	  & $1685\pm648.6$		        \\
 SpaceInvaders	& 				    & $\bmu{845.3\pm69.7}$	& $571.3\pm98.8$ 		    & $\bmu{1706.5\pm118.3}$  & $1648.3\pm294.5$		    \\ \hline

\end{tabular}
\caption{Evaluation scores (mean over 30 evaluation runs with up to 30 initial no-ops) for different training times and algorithms. For each training time limit we compare results from OpenAI ES(our) and Canonical ES $\mu=50$. For each setting we performed 3 training runs, ordered the results for each game and compared them row by row, boldfacing the better score. Results for which the difference is significant across the 30 evaluation runs based on a Mann-Whitney U test ($p<0.05$) are marked in blue.}
\label{table:Scores}
\end{table*}
\vspace*{-0.5cm}
    \paragraph{{Results.}}First, we studied the importance of the parent population size $\mu$. This hyperparameter is known to often be important for ES algorithms and we found this to also hold here. We measured performance for $\mu \in\{10, 20, 50, 100, 200, 400\}$ and observed different optimal values of $\mu$ for different games (Figure \ref{fig:Scores}). For the subsequent analyses we fixed $\mu=50$ for all games.
    
	In Table~\ref{table:Scores}, for each game and for both Canonical ES and OpenAI ES, we report the results of our 3 training runs; for each of these runs, we evaluated the final policy found using 30 rollouts. We ordered the 3 runs according to their mean score and compared the results row-wise. We ran a Mann-Whitney U test~\cite{mann1947test} to check if the distributions of the 30 rewards significantly differed between the two ES variants. After 5 hours of training, Canonical ES performed significantly better than OpenAI ES in 9 out of 24 different runs ($p < 0.05$) and worse in 7 (with 8 ties); this shows that our simple Canonical ES is competitive with the specialized OpenAI ES algorithm on complex benchmark problems from the Atari domain. 
 	Additionally, we tested whether the algorithms still made significant progress between 1 and 5 hours of training; indeed, the performance of our Canonical ES algorithm improved significantly in 16 of 24 cases, demonstrating that it often could make effective use of additional training time.
    However, qualitatively, we observed that the performance of both algorithms tends to plateau in locally optimal solutions for extended periods of time (see Figure \ref{fig:TimePlot}) and that they often find solutions that are not robust to the noise in the domain; i.e., there is a high variance in the points scored with the same trained policy network across initial environment conditions.

\section{Qualitative analysis}\label{Qualitative analysis}
Visual inspection and comparison between solutions found by reinforcement learning algorithms and solutions found by ES algorithms shows some significant differences.
In this section we describe interesting agent behaviors on different games; a video with evaluation runs on all games is available on-line at \url{https://www.youtube.com/watch?v=0wDzPBiURSI}.

\begin{figure}[t]
\centering
\includegraphics[width=0.45\textwidth]{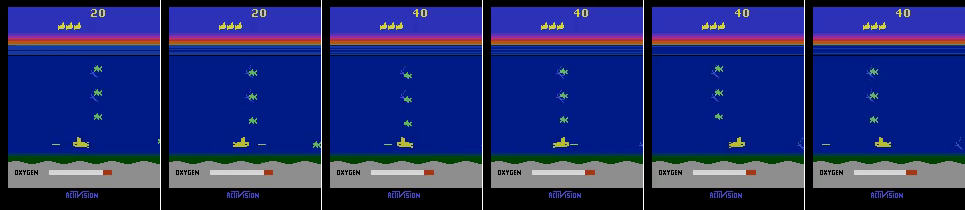}
\vspace*{-0.2cm}
\caption{\label{fig:SeaquestBehaviour}{The agent learns to dive to the bottom of the sea and constantly shoot left and right, occasionally scoring points.}}
\end{figure}

First, we study two games in which most of the ES runs converged to similar sub-optimal solutions: Seaquest and Enduro. In Seaquest, the agent dives to the bottom of the sea and starts to shoot left and right, occasionally hitting an enemy and scoring points (Figure \ref{fig:SeaquestBehaviour}). However, it is not able to detect the lack of oxygen and quickly dies. In Enduro, the agent steers the car to keep it in the middle of the road and accelerate, but from time to time it bounces back after hitting a rival car in front of it. Both solutions are easy-to-reach local optima and do not require developing any complex behavior; since the corresponding scores achieved with these policies are much higher than those of random policies we believe that it is hard for ES to escape from these local attractor states in policy space.

\begin{figure}[t]
\centering
\includegraphics[width=0.45\textwidth]{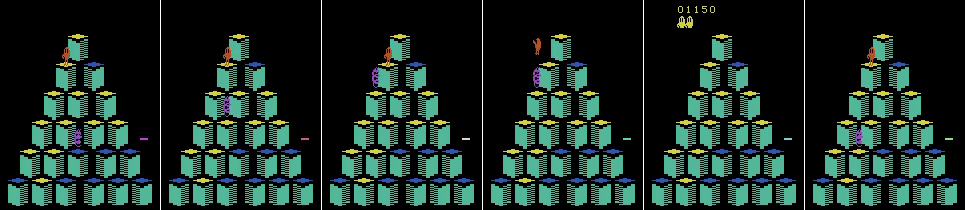}
\vspace*{-0.2cm}
\caption{\label{fig:QbertBehaviourSuicide}{The agent (orange blob in the upper left part of the screen) learns to commit suicide to kill its enemy (purple spring) and collects enough points to get another life. The whole cycle is repeated over and over again.}}
\end{figure}

We next study the game Qbert, in which Canonical ES found two particularly interesting solutions. In the first case (\url{https://www.youtube.com/watch?v=-p7VhdTXA0k}), the agent gathers some points at the beginning of the game and then stops showing interest in completing the level. Instead, it starts to bait an enemy that follows it to kill itself. Specifically, the agent learns that it can jump off the platform when the enemy is right next to it, because the enemy will follow: although the agent loses a life, killing the enemy yields enough points to gain an extra life again (Figure \ref{fig:QbertBehaviourSuicide}). The agent repeats this cycle of suicide and killing the opponent over and over again. 

In the second interesting solution (\url{https://www.youtube.com/watch?v=meE5aaRJ0Zs}), the agent discovers an in-game bug. First, it completes the first level and then starts to jump from platform to platform in what seems to be a random manner. For a reason unknown to us, the game does not advance to the second round but the platforms start to blink and the agent quickly gains a huge amount of points (close to 1 million for our episode time limit). Interestingly, the policy network is not always able to exploit this in-game bug and 22/30 of the evaluation runs (same network weights but different initial environment conditions) yield a low score. 

\begin{figure}[t]
\centering
\includegraphics[width=0.45\textwidth]{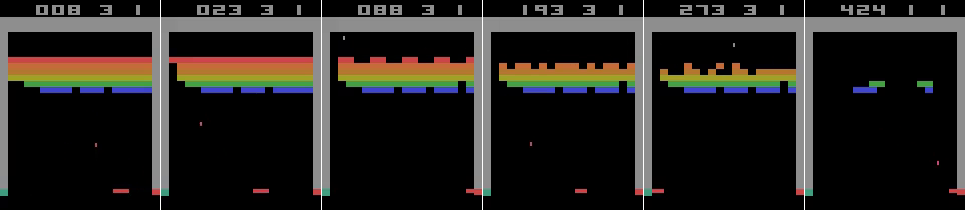}
\vspace*{-0.2cm}
\caption{\label{fig:BreakoutBehaviour}{The agent learns to make a hole in the brick wall to collect many points with one ball bounce.}
}
\end{figure}

Breakout seems to be a challenging environment for ES algorithms. Canonical ES only found reasonable solutions for a few settings. The best solution shows a strategy that looks similar to the best strategies obtained by using reinforcement learning algorithms, in which the agent creates a hole on one side of the board and shoots the ball through it to gain many points in a short amount of time (Figure \ref{fig:BreakoutBehaviour}). However, even this best solution found by ES is not stable: for different initial environment conditions the agent with the same policy network quickly loses the game with only few points. 

In the games SpaceInvaders and Alien we also observed interesting strategies. We do not clip rewards during the training as is sometimes done for reinforcement learning algorithms. Because of that the agent puts more attention to behaviors that result in a higher reward, sometimes even at the cost of the main game objective. In SpaceInvaders, we observe that in the best solution the agent hits the mother-ship that appears periodically on the top of the screen with 100\% accuracy. In Alien, the agent focuses on capturing an item that makes it invincible and then goes to the enemy spawn point to collect rewards for eliminating newly appearing enemies. However, the agent is not able to detect when the invincibility period ends.

\section{Recent related work}

Evolutionary strategies, such as the Covariance Matrix Adaptation Evolution Strategy (CMA-ES~\cite{hansen2003reducing}), are commonly used as a baseline approach in reinforcement learning tasks~ \cite{heidrich2009hoeffding,stulp2013robot,duan2016benchmarking,li2017deep}. Here, we only discuss the most recent related works, which also followed up on the work by \citet{salimans2017evolution}; in particular, we discuss three related arXiv preprints that scientists at Uber released in the last two months about work concurrent to ours.

Similarly to our work, \citet{such2017deep} studied the performance of simpler algorithms than OpenAI's specialized ES variant. They show that genetic algorithms (another broad class of black-box optimization algorithms) can also reach results competitive to OpenAI's ES variant and other RL algorithms. Additionally, interestingly, the authors show that for some of the games even simple random search can outperform carefully designed RL and ES algorithms. 

\citet{lehman2017more} argue that comparing ES to finite-difference-based approximation is too simplistic. The main difference comes from the fact that ES tries to optimize the performance of the distribution of solutions rather than a single solution, thus finding solutions that are more robust to the noise in the parameter space. The authors leave open the question whether this robustness in the parameter space also affects the robustness to the noise in the domain. In our experiments we observe that even for the best solutions on some of the games, the learned policy network is not robust against environment noise. 

\citet{conti2017improving} try to address the problems of local minima (which we also observed in games like Seaquest and Enduro) by augmenting ES algorithms with a novelty search (NS) and quality diversity (QD). Their proposed algorithms add an additional criterion to the optimization procedure that encourages exploring qualitatively different solutions during training, thus reducing the risk of getting stuck in a local optimum. The authors also propose to manage a meta-population of diverse solutions allocating resources to train more promising ones. In our experiments we observe that training runs with the same hyperparameters and different initializations often converge to achieve very different scores; managing a meta-population could therefore be an easy way to improve the results and reduce the variance between the runs. 
Overall, the success of \citet{conti2017improving} in improving performance with some of these newer methods in ES research strengthens our expecation that a wide range of improvements to the state of the art are possible by integrating the multitude of techniques developed in ES research over the last decades into our canonical ES.

\section{Conclusion}

The recent results provided by Open AI~\cite{salimans2017evolution} suggest that natural evolution strategies represent a viable alternative to more common approaches used for deep reinforcement learning. In this work, we analyzed whether the results demonstrated in that work are due to the special type of evolution strategy used there. Our results suggest that even a very basic decades-old evolution strategy 
provides comparable results; thus, more modern evolution strategies should be considered as a potentially competitive approach to modern deep reinforcement learning algorithms.
Since evolution strategies have different strengths and weaknesses than traditional deep reinforcement learning strategies, we also expect rich opportunities for combining the strength of both.

\section*{Acknowledgments}
This work has partly been supported by the European Research Council (ERC) under the European Union's Horizon 2020 research and innovation programme under grant no.\ 716721.
The authors acknowledge support by the state of Baden-Württemberg through bwHPC and the German Research Foundation (DFG) through grant no INST 39/963-1 FUGG. 

\appendix

\footnotesize
\bibliographystyle{named}

\bibliography{ijcai18}

\begin{thebibliography}{}

\bibitem[\protect\citeauthoryear{Brockhoff \bgroup \em et al.\egroup
  }{2010}]{brockhoff2010mirrored}
Dimo Brockhoff, Anne Auger, Nikolaus Hansen, Dirk~V Arnold, and Tim Hohm.
\newblock Mirrored sampling and sequential selection for evolution strategies.
\newblock In {\em International Conference on Parallel Problem Solving from
  Nature}, pages 11--21. Springer, 2010.

\bibitem[\protect\citeauthoryear{Brockman \bgroup \em et al.\egroup
  }{2016}]{brockman2016openai}
Greg Brockman, Vicki Cheung, Ludwig Pettersson, Jonas Schneider, John Schulman,
  Jie Tang, and Wojciech Zaremba.
\newblock Openai gym.
\newblock {\em arXiv preprint arXiv:1606.01540}, 2016.

\bibitem[\protect\citeauthoryear{Clevert \bgroup \em et al.\egroup
  }{2015}]{clevert2015fast}
Djork-Arn{\'e} Clevert, Thomas Unterthiner, and Sepp Hochreiter.
\newblock Fast and accurate deep network learning by exponential linear units
  (elus).
\newblock {\em arXiv preprint arXiv:1511.07289}, 2015.

\bibitem[\protect\citeauthoryear{Conti \bgroup \em et al.\egroup
  }{2017}]{conti2017improving}
Edoardo Conti, Vashisht Madhavan, Felipe~Petroski Such, Joel Lehman, Kenneth~O
  Stanley, and Jeff Clune.
\newblock Improving exploration in evolution strategies for deep reinforcement
  learning via a population of novelty-seeking agents.
\newblock {\em arXiv preprint arXiv:1712.06560}, 2017.

\bibitem[\protect\citeauthoryear{Dhariwal \bgroup \em et al.\egroup
  }{2017}]{baselines}
Prafulla Dhariwal, Christopher Hesse, Matthias Plappert, Alec Radford, John
  Schulman, Szymon Sidor, and Yuhuai Wu.
\newblock Openai baselines.
\newblock \url{https://github.com/openai/baselines}, 2017.

\bibitem[\protect\citeauthoryear{Duan \bgroup \em et al.\egroup
  }{2016}]{duan2016benchmarking}
Yan Duan, Xi~Chen, Rein Houthooft, John Schulman, and Pieter Abbeel.
\newblock Benchmarking deep reinforcement learning for continuous control.
\newblock In {\em ICML}, pages 1329--1338, 2016.

\bibitem[\protect\citeauthoryear{Glasmachers and
  Igel}{2008}]{glasmachers2008uncertainty}
Tobias Glasmachers and Christian Igel.
\newblock Uncertainty handling in model selection for support vector machines.
\newblock In {\em Parallel Problem Solving from Nature (PPSN)}, pages 185--194.
  Springer, 2008.

\bibitem[\protect\citeauthoryear{Gomez \bgroup \em et al.\egroup
  }{2008}]{gomez2008accelerated}
Faustino Gomez, J{\"u}rgen Schmidhuber, and Risto Miikkulainen.
\newblock Accelerated neural evolution through cooperatively coevolved
  synapses.
\newblock {\em Journal of Machine Learning Research}, 9(May):937--965, 2008.

\bibitem[\protect\citeauthoryear{Hansen and
  Ostermeier}{1996}]{hansen1996adapting}
Nikolaus Hansen and Andreas Ostermeier.
\newblock Adapting arbitrary normal mutation distributions in evolution
  strategies: The covariance matrix adaptation.
\newblock In {\em Evolutionary Computation, 1996., Proceedings of IEEE
  International Conference on}, pages 312--317. IEEE, 1996.

\bibitem[\protect\citeauthoryear{Hansen \bgroup \em et al.\egroup
  }{2003}]{hansen2003reducing}
Nikolaus Hansen, Sibylle~D M{\"u}ller, and Petros Koumoutsakos.
\newblock Reducing the time complexity of the derandomized evolution strategy
  with covariance matrix adaptation (cma-es).
\newblock {\em Evolutionary computation}, 11(1):1--18, 2003.

\bibitem[\protect\citeauthoryear{Heidrich-Meisner and
  Igel}{2009}]{heidrich2009hoeffding}
Verena Heidrich-Meisner and Christian Igel.
\newblock Hoeffding and bernstein races for selecting policies in evolutionary
  direct policy search.
\newblock In {\em Proceedings of the 26th Annual International Conference on
  Machine Learning}, pages 401--408. ACM, 2009.

\bibitem[\protect\citeauthoryear{Igel}{2011}]{igel2011evolutionary}
Christian Igel.
\newblock Evolutionary kernel learning.
\newblock In {\em Encyclopedia of Machine Learning}, pages 369--373. Springer,
  2011.

\bibitem[\protect\citeauthoryear{Ioffe and Szegedy}{2015}]{ioffe2015batch}
Sergey Ioffe and Christian Szegedy.
\newblock Batch normalization: Accelerating deep network training by reducing
  internal covariate shift.
\newblock In {\em ICML}, pages 448--456, 2015.

\bibitem[\protect\citeauthoryear{Kingma and Ba}{2014}]{kingma2014adam}
Diederik~P Kingma and Jimmy Ba.
\newblock Adam: A method for stochastic optimization.
\newblock {\em arXiv preprint arXiv:1412.6980}, 2014.

\bibitem[\protect\citeauthoryear{Lehman \bgroup \em et al.\egroup
  }{2017}]{lehman2017more}
Joel Lehman, Jay Chen, Jeff Clune, and Kenneth~O Stanley.
\newblock Es is more than just a traditional finite-difference approximator.
\newblock {\em arXiv preprint arXiv:1712.06568}, 2017.

\bibitem[\protect\citeauthoryear{Li}{2017}]{li2017deep}
Yuxi Li.
\newblock Deep reinforcement learning: An overview.
\newblock {\em arXiv preprint arXiv:1701.07274}, 2017.

\bibitem[\protect\citeauthoryear{Loshchilov and
  Hutter}{2016}]{loshchilov2016cma}
Ilya Loshchilov and Frank Hutter.
\newblock {CMA-ES for Hyperparameter Optimization of Deep Neural Networks}.
\newblock {\em arXiv preprint arXiv:1604.07269}, 2016.

\bibitem[\protect\citeauthoryear{Mann and Whitney}{1947}]{mann1947test}
Henry~B Mann and Donald~R Whitney.
\newblock On a test of whether one of two random variables is stochastically
  larger than the other.
\newblock {\em The annals of mathematical statistics}, pages 50--60, 1947.

\bibitem[\protect\citeauthoryear{Mnih \bgroup \em et al.\egroup
  }{2013}]{mnih2013playing}
Volodymyr Mnih, Koray Kavukcuoglu, David Silver, Alex Graves, Ioannis
  Antonoglou, Daan Wierstra, and Martin Riedmiller.
\newblock Playing atari with deep reinforcement learning.
\newblock {\em arXiv preprint arXiv:1312.5602}, 2013.

\bibitem[\protect\citeauthoryear{Mnih \bgroup \em et al.\egroup
  }{2015}]{mnih2015human}
Volodymyr Mnih, Koray Kavukcuoglu, David Silver, Andrei~A Rusu, Joel Veness,
  Marc~G Bellemare, Alex Graves, Martin Riedmiller, Andreas~K Fidjeland, Georg
  Ostrovski, et~al.
\newblock Human-level control through deep reinforcement learning.
\newblock {\em Nature}, 518(7540):529--533, 2015.

\bibitem[\protect\citeauthoryear{Mnih \bgroup \em et al.\egroup
  }{2016}]{mnih2016asynchronous}
Volodymyr Mnih, Adria~Puigdomenech Badia, Mehdi Mirza, Alex Graves, Timothy
  Lillicrap, Tim Harley, David Silver, and Koray Kavukcuoglu.
\newblock Asynchronous methods for deep reinforcement learning.
\newblock In {\em ICML}, pages 1928--1937, 2016.

\bibitem[\protect\citeauthoryear{Rechenberg}{1973}]{rechenberg1973evolutionsstrategie}
Ingo Rechenberg.
\newblock Evolutionsstrategie--optimierung technisher systeme nach prinzipien
  der biologischen evolution.
\newblock 1973.

\bibitem[\protect\citeauthoryear{Rudolph}{1997}]{rudolph1997convergence}
G{\"u}nter Rudolph.
\newblock {\em Convergence properties of evolutionary algorithms}.
\newblock Kovac, 1997.

\bibitem[\protect\citeauthoryear{Salimans \bgroup \em et al.\egroup
  }{2016}]{salimans2016improved}
Tim Salimans, Ian Goodfellow, Wojciech Zaremba, Vicki Cheung, Alec Radford, and
  Xi~Chen.
\newblock Improved techniques for training gans.
\newblock In {\em Advances in Neural Information Processing Systems}, pages
  2234--2242, 2016.

\bibitem[\protect\citeauthoryear{Salimans \bgroup \em et al.\egroup
  }{2017}]{salimans2017evolution}
Tim Salimans, Jonathan Ho, Xi~Chen, and Ilya Sutskever.
\newblock Evolution strategies as a scalable alternative to reinforcement
  learning.
\newblock {\em arXiv preprint arXiv:1703.03864}, 2017.

\bibitem[\protect\citeauthoryear{Stulp and Sigaud}{2013}]{stulp2013robot}
Freek Stulp and Olivier Sigaud.
\newblock Robot skill learning: From reinforcement learning to evolution
  strategies.
\newblock {\em Paladyn, Journal of Behavioral Robotics}, 4(1):49--61, 2013.

\bibitem[\protect\citeauthoryear{Such \bgroup \em et al.\egroup
  }{2017}]{such2017deep}
Felipe~Petroski Such, Vashisht Madhavan, Edoardo Conti, Joel Lehman, Kenneth~O
  Stanley, and Jeff Clune.
\newblock Deep neuroevolution: Genetic algorithms are a competitive alternative
  for training deep neural networks for reinforcement learning.
\newblock {\em arXiv preprint arXiv:1712.06567}, 2017.

\bibitem[\protect\citeauthoryear{Todorov \bgroup \em et al.\egroup
  }{2012}]{todorov2012mujoco}
Emanuel Todorov, Tom Erez, and Yuval Tassa.
\newblock Mujoco: A physics engine for model-based control.
\newblock In {\em IROS, 2012 IEEE/RSJ International Conference on}, pages
  5026--5033. IEEE, 2012.

\bibitem[\protect\citeauthoryear{Wierstra \bgroup \em et al.\egroup
  }{2008}]{wierstra2008natural}
Daan Wierstra, Tom Schaul, Jan Peters, and Juergen Schmidhuber.
\newblock Natural evolution strategies.
\newblock In {\em Evolutionary Computation, 2008. CEC 2008.(IEEE World Congress
  on Computational Intelligence). IEEE Congress on}, pages 3381--3387. IEEE,
  2008.

\bibitem[\protect\citeauthoryear{Wierstra \bgroup \em et al.\egroup
  }{2014}]{wierstra2014natural}
Daan Wierstra, Tom Schaul, Tobias Glasmachers, Yi~Sun, Jan Peters, and
  J{\"u}rgen Schmidhuber.
\newblock Natural evolution strategies.
\newblock {\em Journal of Machine Learning Research}, 15(1):949--980, 2014.

\bibitem[\protect\citeauthoryear{Williams}{1992}]{williams1992simple}
Ronald~J Williams.
\newblock Simple statistical gradient-following algorithms for connectionist
  reinforcement learning.
\newblock In {\em Reinforcement Learning}, pages 5--32. Springer, 1992.

\end{thebibliography}

\end{document}